\pdfoutput=1
\documentclass[11pt]{article}

\usepackage[preprint]{acl}
\usepackage{float}
\usepackage{times}
\usepackage{latexsym}
\usepackage{xcolor}

\usepackage[T1]{fontenc}
\usepackage[utf8]{inputenc}

\usepackage{microtype}

\usepackage{inconsolata}

\usepackage{graphicx}

\usepackage{times}
\usepackage{latexsym}

\usepackage{graphics}
\usepackage{graphicx}
\usepackage{color}
\usepackage{colortbl}
\usepackage{amsfonts}
\usepackage{amssymb}
\usepackage{amsmath}
\usepackage{nicefrac}
\usepackage{bbm}
\usepackage{booktabs}
\usepackage{microtype}
\usepackage{multirow}
\usepackage{multicol}
\usepackage{tabu}
\usepackage{amsthm}
\usepackage{footmisc}
\usepackage{soul}
\usepackage{arydshln}
\usepackage{wasysym}
\usepackage{array}
\usepackage{makecell}
\usepackage{scalerel}

\usepackage{algpseudocode}
\usepackage{algorithm}
\usepackage{hyperref}
\usepackage{xspace}
\usepackage{url}
\usepackage{utfsym}
\usepackage{dirtytalk}

\usepackage{float}
\usepackage{enumitem}
\usepackage{tikz}
\usepackage{tcolorbox}
\usepackage{siunitx}

\usepackage{graphicx} 
\usepackage{subcaption} %  for subfigures environments 

\definecolor{lightgrey}{RGB}{242, 242, 242}

\newlength\mylen
\newtcolorbox[auto counter]{promptfloat}[2][]{%
    float=!t,%
    blend before title=dash hang,%
    title={\textbf{Prompt~\thetcbcounter:} #2},%
    colback=black!5!white,%
    colframe=black!70!white,%
    #1}

\newcommand{\prompt}[3]{%
    \begin{promptfloat}[label={#1}]{#2}
    {\vspace{12pt}\small \texttt{#3}\vspace{12pt}}
\end{promptfloat}
\vspace{5pt}
}

\title{Meta-Reasoning Improves Tool Use in Large Language Models}

\author{Lisa Alazraki \\
  Imperial College London \\
  \texttt{lisa.alazraki20@imperial.ac.uk} \\\And
  Marek Rei \\
  Imperial College London \\
  \texttt{marek.rei@imperial.ac.uk} \\}

\begin{document}
\maketitle

\begin{abstract}
External tools help large language models succeed at tasks where they would otherwise typically fail. In existing frameworks, choosing tools at test time relies on naive greedy decoding, regardless of whether the model has been fine-tuned on tool-annotated data or prompted with in-context examples. In contrast, we find that gathering and choosing among a suitable set of candidate tools has greater potential to lead to an optimal selection. We present \textbf{T}ool sel\textbf{\textsc{ect}}ion via meta-reas\textbf{\textsc{on}}ing (\textsc{Tecton}), a two-phase system that first \textit{reasons} over a task and outputs candidate tools using a custom fine-tuned language modelling head. Then, with the custom head disabled, it \textit{meta-reasons} (i.e., it reasons over the previous reasoning process) to make a final choice.  We show that \textsc{Tecton} results in substantial gains—both in-distribution and out-of-distribution—on a range of math reasoning datasets.
\end{abstract}
\vspace{5pt}

\section{Introduction}

\begin{figure*}
\includegraphics[width=\linewidth]{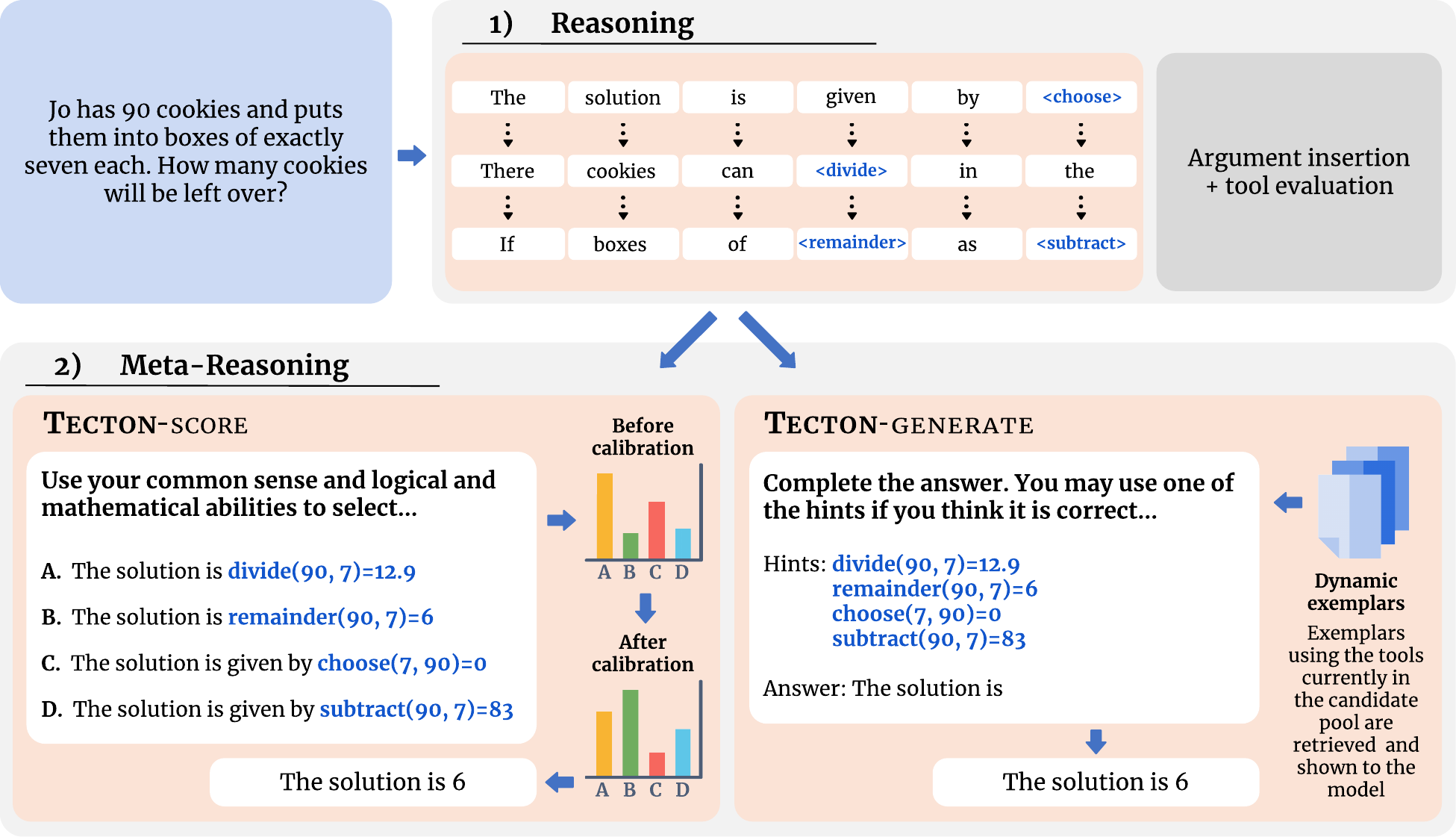}

\caption{An overview of \textsc{Tecton}. In the reasoning phase, the system inspects the task and decodes a set of candidate tools, followed by argument insertion and evaluation of the tools via the Python interpreter. In the meta-reasoning phase, the model is asked to select the most useful tool, either by scoring multiple options (\textsc{Tecton-score}) or by continuing the generation given the decoded tools as hints (\textsc{Tecton-generate}).}
\label{fig:system}
\end{figure*}

Augmentation with external tools has proven effective at boosting the performance of large language models (LLMs) in knowledge-intensive tasks such as QA and math problem-solving \cite{NEURIPS2023_8fd1a81c, paranjape2023art, parisi2022talm, NEURIPS2023_d842425e}. Tools are self-contained programs or APIs which the model can execute with chosen arguments as needed. To teach an LLM how to use tools, previous work adopts one of three main strategies: (1) tool demonstrations via in-context learning (ICL) \cite{10.5555/3618408.3618843, 10204174, hsieh2023tool, surismenon2023vipergpt}, (2) full model fine-tuning on a dataset where text samples are interleaved with tool annotations \cite{parisi2022talm, NEURIPS2023_d842425e, tang2023toolalpaca, patil2023gorilla}, or (3) parameter-efficient fine-tuning (PEFT) on tool annotated data \cite{NEURIPS2023_8fd1a81c, qiao2024trice, wang2024guiding}. Similarly to full fine-tuning, PEFT methods can teach LLMs a very large number of tools, while ICL is limited by the fixed size of the context window \cite{NEURIPS2023_8fd1a81c, patil2023gorilla}. Although the fine-tuning paradigm in general binds the model to the set of tools learned during training, parameter-efficient learning reduces the cost of tuning, thus facilitating potential future extensions of the tool set. In contrast, adding further tools via a new round of full-model tuning incurs a significant computational cost \cite{NEURIPS2023_8fd1a81c}. We note that a further advantage of PEFT is that it tunes a handful of additional task-specific parameters, which can be selectively disabled to reinstate the frozen model and its original capabilities \cite{ding23naturemi, han2024parameterefficient}. 

In existing literature, inference-time tool selection is made by greedily decoding the most likely tool \cite{10.5555/3618408.3618843, NEURIPS2023_8fd1a81c, NEURIPS2023_d842425e, wang2024guiding}, regardless of whether the model has been fully fine-tuned, PEFT-tuned, or prompted in-context. In this work, we revisit the paradigm that selects tools solely based on their probability at decoding time. We propose an alternative, novel framework for \textbf{T}ool sel\textbf{\textsc{ect}}ion via meta-reas\textbf{\textsc{on}}ing (\textsc{Tecton}) that gathers and chooses among a suitable set of candidate tools. We train a parameter-efficient language modelling head on tool-annotated data, similar to \citet{NEURIPS2023_8fd1a81c}, which can be switched on or off as needed. \textsc{Tecton} thus comprises two distinct phases: in the \textit{reasoning phase}, it investigates a task and outputs candidate tools with the aid of the tuned LM head. Then, in the \textit{meta-reasoning phase}, it uses the frozen LLM to re-examine the candidates and make a final decision. Fig.~\ref{fig:system} illustrates the framework. We train and evaluate \textsc{Tecton} on math reasoning datasets, following established work on LLM tool calling \cite{chen-etal-2024-good, das-etal-2024-mathsensei, gou2024toratoolintegratedreasoningagent}. Among tasks that benefit from tools, math reasoning is particularly challenging, since it requires chains of multiple tool calls with errors that compound. This is evidenced by existing tool-augmented LLMs, which achieve the lowest performance on math reasoning when evaluated on multiple tasks \cite{NEURIPS2023_8fd1a81c, NEURIPS2023_d842425e}.
In summary, our main contributions are:\vspace{-5pt}
\begin{itemize}
    \item We introduce \textsc{Tecton}, a novel two-phase framework that combines a custom fine-tuned head with a frozen LLM to improve tool use (Section \ref{sec:method}).
    \item We show that \textsc{Tecton} outperforms strong baselines in math reasoning tasks, both on in-distribution data and on unseen benchmarks (Section \ref{sec:experiments}).
    \item We enhance three popular math reasoning datasets to make them more challenging for current LLMs. We share our data and code at 
    \url{https://github.com/lisaalaz/tecton}. 
\end{itemize}

\section{Method}
\label{sec:method}
\subsection{Preliminaries}
We augment the language modelling head of a base model with additional token embeddings $T$ to represent math operations, and train them via a standard language modelling objective. Once trained, $T$ comprises the tools available to the LLM for solving math problems. Prior work has shown the effectiveness of tuning additional tokens for both math tasks \cite{NEURIPS2023_8fd1a81c, wang2024guiding} and general reasoning \cite{goyal2024thinkspeaktraininglanguage, herel2024thinkingtokenslanguagemodeling}. 

Our preliminary experiments show that in cases where the LLM has failed to generate the correct tool by greedy sampling, this can usually be found among tokens that have only slightly lower probabilities (Figure~\ref{fig:lineplot}). Over-sampling an appropriate set of candidate tools may thus be a better strategy than greedily decoding the most likely tool, provided this is combined with a reliable method for choosing among the candidates. To this end, we design a two-phase framework that leverages both the specialised, augmented LM head (\textit{reasoning phase}) and the underlying generalist LLM (\textit{meta-reasoning phase}). The rest of the section describes this in detail.

\subsection{Reasoning Phase}\label{sec:reasoning}
Given a math problem where individual reasoning steps are separated by newline tokens, we ask the LLM to solve it line by line, and collect a set of candidate tools for each line using the augmented LM head. We experiment with both temperature sampling and greedy decoding of multiple tokens (see Appendix \ref{sec:sampling}). We find that looking at the top $k$ most likely tokens at every position in the sequence is the most promising approach to gather a diverse yet relevant pool of candidates. Hence, we generate an intermediate solution to each line of the problem, gathering tools from the top $k$ tokens at each decoding step. For each line, the multiset $C$ of candidate tools is given by $C = \{\!\{ W_{ij} \in T, {1 \leq i \leq l}, \, { 1 \leq j \leq k} \}\!\}$,
where $T$ is the set of available tool tokens, and $W$ is the matrix resulting from decoding $k$ top-probability tokens at each of the $l$ token positions in the line of text. We set $k=5$ as a trade-off between search space size and computation cost. We then prompt the LLM to produce arguments for each candidate tool given the previous context. Identical tools with the same arguments are dropped from the pool. Finally, we pass each candidate tool and arguments into the Python interpreter, and keep those that are successfully evaluated.
Once a line of text toward the solution has been processed in this way, we move onto the meta-reasoning phase.

\subsection{Meta-Reasoning Phase}

In this phase, we disable the custom-tuned head and let the underlying LLM analyse its previous reasoning process to choose among the candidate tools.  Frozen LLMs have been used to self-evaluate and meta-reason over previous answers in existing literature \cite{pmlr-v239-alazraki23a, NEURIPS2023_1b44b878, NEURIPS2023_271db992, zeng2024mrgsm8kmetareasoningbenchmarklarge}. 
We experiment with two ways of eliciting meta-reasoning: \textsc{Tecton}-\textsc{score} and \textsc{Tecton}-\textsc{generate}.
\paragraph{\textsc{Tecton}-\textsc{score}.} We join each candidate tool with the previous context, and present these as options for the LLM to \textit{score}. We prefix each option with an uppercase letter label and select as the answer continuation the option whose label is assigned highest probability by the model, i.e., $\text{arg max}_{t_i \in \mathcal{V}_{\text{sub}}} \, p(t_i \mid t_{{}<i} \text{ {with} } t_{{}<i} \in \mathcal{V})$, where $\mathcal{V}_{\text{sub}}$ denotes a subset of the vocabulary containing only the uppercase letter tokens that are in the label set. In this setup, 
% which we call \textsc{Tecton}-\textsc{score},
we limit the number of candidates to a maximum of four.
\paragraph{\textsc{Tecton}-\textsc{generate}.} We pass the candidate tools as hints and ask the model to \textit{generate} an appropriate continuation of the answer. Here, the hints serve as mere guidance for the LLM (i.e., the model could choose to ignore all candidates and generate something different). In this version of the system,
% which we call \textsc{Tecton-generate}, 
the model benefits from dynamically retrieved few-shot exemplars demonstrating the tools in the candidate set.

\begin{figure}[t]
    \includegraphics[width=\linewidth]{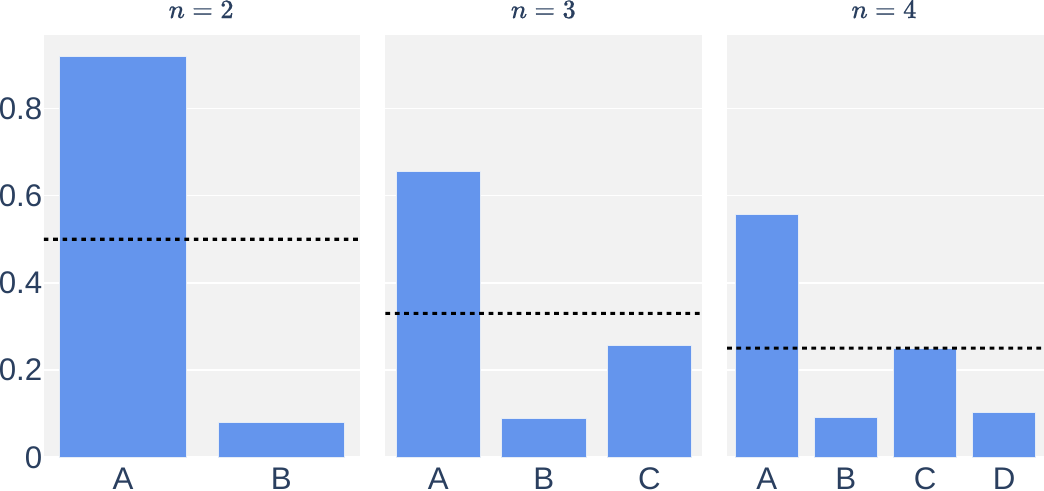}
    \caption{Averaged biased probability distributions over $n$ labels (for $n=\lbrace1, 2, 3\rbrace$), obtained on GSM8K-XL's validation set. The dotted lines indicate the uniform averaged distribution that would be given by an unbiased model. Note that the label distribution is similarly skewed for FuncQA.}
    \label{fig:biases}

\end{figure}

\subsection{Bias Calibration} 
Upon running \textsc{Tecton}-\textsc{score} without recalibration of the label probabilities, we find that the validation results are poor. Visual inspection of the samples reveals that the model assigns highest likelihood to the same label in most instances, as shown in Fig. \ref{fig:biases}. This is consistent with \citet{llm-mcq-bias}'s finding that LLMs are prone to selection bias in multiple-choice tasks. To solve a similar problem, \citet{duarte2024decop} measure their model's bias over the labels $A$, $B$, $C$, $D$ using a set of neutral samples where a uniform distribution would be expected, and subtract that bias from each label's likelihood at inference time. Our math reasoning task does not lend itself to finding neutral samples, so we adopt a different strategy. Having created data samples of questions and options (by running the reasoning phase of \textsc{Tecton} on a validation set), we compute $n!$ permutations of the $n$ options while keeping the letter labels in the same position. We have the model score the labels for each permutation, and average over all permutations and all data samples to obtain an averaged biased distribution $B^{(n)}$. Since the number of options in our task is variable, we run this process independently for data samples with $n=2$, $n=3$ and $n=4$ labels. At inference time, given a set of labels $L^{(n)}$ of size $n$, we retrieve the corresponding $B^{(n)}$ and compute the calibrated probability $\hat{p}_i$ of each label $l_i$ in the set as
\[ \hat{p}_i = p_i + \frac{1}{n} - {B^{(n)}}_i\]
where $p_i$ is the probability assigned by the model to label $l_i$ for the current sample, ${B^{(n)}}_i$ is the pre-computed biased probability of label $l_i$, and $n$ is the total number of labels.

\subsection{Retrieval of Tool Demonstrations}
To aid answer generation in \textsc{Tecton-generate}, we dynamically retrieve and add to the context few-shot exemplars demonstrating the candidate tools. We create the retrieval pool by extracting training samples and collecting candidate tools for each, by running the reasoning phase of \textsc{Tecton}. We construct each exemplar to simulate the inference task, as follows: (1) we append to the sample its set of candidate tools, and (2) we append to it the golden answer demonstrating how the correct tool is used to obtain the final solution. At inference time, we retrieve only the exemplars whose golden answers contain the tools currently in the candidate set. It is worth noting that dynamic retrieval was not included in \textsc{Tecton-score} as it did not improve validation performance.

\section{Experiments}\label{sec:experiments}

\subsection{Experimental Setup}

Our system is model-agnostic and can be applied to any open-weights LLM. Here, we use Llama 3 8B Instruct \cite{dubey2024llama3herdmodels} (henceforth referred to as Llama 3) as the base model in all experiments. Implementation details and hyperparameters are given in Appendix \ref{sec:model}.

\subsection{Datasets}

We train and evaluate \textsc{Tecton} on GSM8K-XL \cite{cobbe2021training, NEURIPS2023_8fd1a81c} and FuncQA \cite{NEURIPS2023_8fd1a81c}. The test set of the latter is comprised of two distinct subsets: a `one-hop' corpus containing problems solvable with one single operation (FuncQA-OH), and a `multi-hop' one requiring multiple operations (FuncQA-MH). For GSM8K-XL we tune four additional tokens corresponding to the four basic operations, and extend these to 13 in the case of FuncQA. The complete set of tools for each dataset is shown in Appendix \ref{sec:model}. Additionally, we evaluate on out-of-distribution datasets that were not observed during fine-tuning. For this purpose we choose a range of math reasoning datasets: ASDiv \cite{miao-etal-2020-diverse}, MAWPS \cite{koncel-kedziorski-etal-2016-mawps} and SVAMP \cite{patel-etal-2021-nlp}. These were found by \citet{hub_nature} to be OOD with respect to GSM8K: not only do they display minimal n-gram overlap, but they also require reasoning chains of different average lengths from GSM8K to be solved. These datasets have been used to evaluate LLMs in previous works, both with and without the aid of tools \cite{NEURIPS2022_8bb0d291, NEURIPS2023_d842425e}. Following a strategy similar to the one used by \citet{NEURIPS2023_8fd1a81c} for constructing GSM8K-XL, we magnify the numbers in these datasets to make them challenging for current LLMs (the enhancement process is described in Appendix \ref{sec:datasets}). We thus obtain ASDiv-XL, MAWPS-XL, and SVAMP-XL. We use these datasets to test models tuned on GSM8K-XL.

\subsection{Baselines}

We implement recent tool-augmented models as baselines: \textsc{Trice} \cite{qiao2024trice} and ToolkenGPT \cite{NEURIPS2023_8fd1a81c}. These share a parameter-efficient approach with \textsc{Tecton} (see Appendix~\ref{sec:baselines} for details). Despite their relatively low computational cost, they have been shown to outperform strong systems: \textsc{Trice} paired with Alpaca \cite{alpacareport}, ChatGLM \cite{glm2024chatglmfamilylargelanguage} and Vicuna \cite{vicunareport} surpasses the much larger GPT-3.5 as well as tool learning via supervised fine-tuning. In addition to outperforming GPT-3.5, ToolkenGPT paired with LLaMA \cite{touvron2023llama} is more accurate than ReAct \cite{react}. It should also be noted that LLMs are increasingly able to solve math problems and perform difficult arithmetic without tools, as shown by the high accuracy ($79.6\%$) achieved by Llama 3 on the non-enhanced version of GSM8K \cite{dubey2024llama3herdmodels}. Hence we additionally compare against a vanilla version of Llama 3 as well as Chain-of-Thought (CoT) prompting \cite{NEURIPS2022_9d560961} with exemplars extracted from the train set.

\begin{table*}[t!]
\centering
\renewcommand\arraystretch{1.35}
\setlength{\tabcolsep}{12pt}
\scalebox{0.68}{
\begin{tabular}{lcccccc}
\hline
\toprule
\multicolumn{1}{c}{} 
& \multicolumn{3}{c}{\textbf{\Large{In-distribution}}} 
& \multicolumn{3}{c}{\textbf{\Large{Out-of-distribution}}}
\\
\cmidrule(lr){2-4} \cmidrule(lr){5-7} 
& \textbf{FuncQA-OH} & \textbf{FuncQA-MH} & \textbf{GSM8K-XL} 
% & \LG \large{\textbf{Avg.}}  
& \textbf{ASDiv-XL} & \textbf{MAWPS-XL} & \textbf{SVAMP-XL}
% & \LG \large{\textbf{Avg.}} 
\\
\Xhline{1px}

 \Large{Llama 3} 
& \Large{10.0} & \Large{2.9}  & \Large{13.0}  
% & \LG \Large{-} 
& \Large{25.8}  & \Large{26.2} & \Large{27.8}  
% & \LG \Large{-}  
\\

\small{$+$ }\Large{ CoT } 
& \Large{25.0} & \Large{5.9}  & \Large{37.3}  
% & \LG \Large{22.7} 
& \Large{40.6}  & \Large{58.7} & \Large{51.9}  
% & \LG \Large{50.4}  
\\

\Xhline{0.5px}

\small{$+$ }\Large{ \textsc{Trice}} 
& \Large{-}  & \Large{-}  & \Large{43.5} 
% & \LG \Large{-}
& \Large{52.2}  & \Large{71.2} & \Large{49.6} 
% & \LG \Large{46.5}  
\\

\small{$+$ }\Large{ ToolkenGPT} 
& \Large{65.0}  & \Large{10.3}  & \Large{48.4} 
% & \LG \Large{41.2}
& \Large{45.3}  & \Large{68.3} & \Large{60.4} 
% & \LG \Large{58.0}  
\\

\Xhline{0.5px}

{\small{$+$ }\Large{ \textsc{Tecton-score}}}
& \Large{66.7}  & \Large{17.6}  & \textbf{\Large{50.7}} 
% & \LG \Large{45.0}
& \Large{53.6}  & \Large{76.9} & \Large{62.2}  
% & \LG \Large{64.2}  
\\

{\small{$+$ }\Large{ \textsc{Tecton-generate}}}
& \textbf{\Large{70.0}}  & \textbf{\Large{20.6}}  & \Large{45.8} 
% & \LG \Large{\textbf{45.5}}
& \textbf{\Large{55.3}} & \textbf{\Large{77.4}} & \textbf{\Large{66.7}} 
% & \LG \Large{\textbf{66.5}}  
\\

\bottomrule
\hline
\end{tabular}
}

\caption{
Accuracies on math reasoning datasets measured via exact match of the result rounded to 2 decimal places. We do not tune \textsc{Trice} on FuncQA as this dataset lacks the necessary annotations for RLEF training. Note that the separation between in-distribution and out-of-distribution data does not apply to the vanilla version of Llama 3.
}
\label{tab:main_result}
\end{table*}

\begin{table*}[t!]
\centering
\renewcommand\arraystretch{1.35}
\setlength{\tabcolsep}{12pt}
\scalebox{0.68}{
\begin{tabular}{lcccccc}
\hline
\toprule
\multicolumn{1}{c}{} 
& \multicolumn{3}{c}{\textbf{\Large{In-distribution}}}
& \multicolumn{3}{c}{\textbf{\Large{Out-of-distribution}}}
\\
\cmidrule(lr){2-4} \cmidrule(lr){5-7} 
& \textbf{FuncQA-OH} & \textbf{FuncQA-MH} & \textbf{GSM8K-XL} & \textbf{ASDiv-XL} & \textbf{MAWPS-XL} & \textbf{SVAMP-XL}
\\
\Xhline{1px}

\makecell[l]{\\[-1.6ex] \Large{\textsc{Tecton-score}} \\ \textbf{-- bias calibration} \\[0.4ex]}
& \Large{58.3}  & \Large{8.8}  & \Large{49.3} 
& \Large{52.2}  & \Large{75.0} & \Large{61.9}   \\

\Xhline{0.5px}

\makecell[l]{\\[-1.6ex] \Large{\textsc{Tecton-generate}} \\ \textbf{-- dynamic exemplar retrieval} \\[0.4ex]}
& \Large{58.3}  & \Large{13.2}  & \Large{43.5} 
& \Large{50.8} & \Large{71.2} & \Large{62.9}  \\

\bottomrule
\hline
\end{tabular}
}

\caption{
Results of ablating components of \textsc{Tecton-score} and \textsc{Tecton-generate}, on in-distribution and out-of-distribution data.
}
\label{tab:detailedablation}
\end{table*}

\subsection{Results}

Table \ref{tab:main_result} shows \textsc{Tecton}'s gains on math reasoning. Both versions of the system achieve scores above baselines for in-distribution and out-of-distribution-data, with the exception of \textsc{Tecton-generate} on GSM8K-XL, whose performance is slightly below that of ToolkenGPT.

\paragraph{In-distribution performance.} On GSM8K-XL, our best implementation scores $7.2$ percentage points above \textsc{Trice} and $2.3$ above ToolkenGPT. When evaluated on in-distribution data, \textsc{Tecton}'s most significant gains are on FuncQA: the \textsc{generate} setting doubles the performance of ToolkenGPT in the challenging multi-hop task, reaching $20.6\%$ accuracy. For context, Llama 3 can only solve $2.9\%$ of the dataset and CoT only raises performance to $5.9\%$. 

\paragraph{Out-of-distribution performance.} On OOD datasets, \textsc{Tecton}'s advantage is substantial: on average, \textsc{Tecton-generate} gains $8.5$ percentage points over ToolkenGPT and $8.8$ over \textsc{Trice}. \textsc{Tecton-score} achieves $6.2$ and $6.6$ above the two baselines, respectively. This highlights the adaptability of the method to unseen data.

\subsection{Ablations}

To gain more insight into these results, we perform an ablation study on the meta-reasoning phase of \textsc{Tecton}, shown in Table \ref{tab:detailedablation}. We ablate bias calibration from \textsc{Tecton-score} and dynamic exemplar retrieval from \textsc{Tecton-generate}. On average, \textsc{Tecton-score}'s ablated accuracy is $6.2$ percentage points lower than the non-ablated version on in-distribution data, and $1.2$ on unseen datasets. Additionally, \textsc{Tecton-generate}'s average performance drops by $7.2$ on in-distribution data and $4.9$ on OOD data. The most significant performance loss is on FuncQA: the ablated versions of \textsc{Tecton-score} and \textsc{Tecton-generate} see a decrease of $8.4$ and $11.7$ percentage points, respectively, on the one-hop test set. They also drop by $8.8$ and $7.4$, respectively, on the multi-hop split. On GSM8K, the decline is less severe: $-1.4$ for \textsc{Tecton-score} and $-2.3$ for \textsc{Tecton-generate}. \textsc{Tecton-score}'s decrease after ablation is similarly modest on OOD datasets, while \textsc{Tecton-generate}'s is more significant: its ablated performance drops by $6.2$ points on MAWPS, $4.5$ on ASDiv and $3.8$ on SVAMP. This highlights the benefit of dynamic exemplar retrieval during the meta-reasoning phase of \textsc{Tecton-generate}. Despite the performance decline, we find that both ablated versions of \textsc{Tecton} outperform CoT on all datasets. They also consistently surpass ToolkenGPT's accuracy on OOD datasets, and either match or outperform \textsc{Trice}, with the sole exception of \textsc{Tecton-generate} on ASDiv-XL.

\section{Conclusion}

We introduce \textsc{Tecton}, a novel two-phase framework that first samples a set of candidate tools and then selects the optimal candidate via meta-reasoning. We implement two versions of the system and find that both achieve superior performance on math reasoning datasets, surpassing our strongest baseline by $\sim4\%$ on in-distribution data and $\sim9\%$ on unseen benchmarks, on average. These results confirm our hypothesis that a specialized, custom-tuned framework and a generalist pre-trained model can work together to improve tool use in challenging tasks.

\section*{Limitations}

This paper solely focuses on math reasoning tasks. While this is consistent with established literature, there are other domains (e.g., knowledge-intensive QA, virtual environment navigation) that can benefit from the use of tools. Future work can investigate a wider range of tasks.

\section*{Ethical Considerations}

We have verified that all datasets and software utilized in this paper allow for their use, distribution and modification. Our non-commercial purpose is consistent with all licenses. The distribution of our code and data is accompanied by the licenses and credits to the original authors.

\section*{Acknowledgments}

The authors would like to thank Joe Stacey for his insightful comments on the first draft of this paper.

\bibliography{acl_latex}

\begin{thebibliography}{41}
\providecommand{\natexlab}[1]{#1}

\bibitem[{Alazraki et~al.(2023)Alazraki, Castrejon, Dehghani, Huot, Uijlings, and Mensink}]{pmlr-v239-alazraki23a}
Lisa Alazraki, Lluis Castrejon, Mostafa Dehghani, Fantine Huot, Jasper Uijlings, and Thomas Mensink. 2023.
\newblock \href {https://proceedings.mlr.press/v239/alazraki23a.html} {How (not) to ensemble {LVLMs} for {VQA}}.
\newblock In \emph{Proceedings on "I Can't Believe It's Not Better: Failure Modes in the Age of Foundation Models" at NeurIPS 2023 Workshops}, volume 239 of \emph{Proceedings of Machine Learning Research}, pages 1--20. PMLR.

\bibitem[{Chen et~al.(2024)Chen, Li, Wang, and Li}]{chen-etal-2024-good}
Nuo Chen, Hongguang Li, Baoyuan Wang, and Jia Li. 2024.
\newblock \href {https://aclanthology.org/2024.nlrse-1.7} {From good to great: Improving math reasoning with tool-augmented interleaf prompting}.
\newblock In \emph{Proceedings of the 2nd Workshop on Natural Language Reasoning and Structured Explanations (@ACL 2024)}, pages 64--79, Bangkok, Thailand. Association for Computational Linguistics.

\bibitem[{Cobbe et~al.(2021)Cobbe, Kosaraju, Bavarian, Chen, Jun, Kaiser, Plappert, Tworek, Hilton, Nakano, Hesse, and Schulman}]{cobbe2021training}
Karl Cobbe, Vineet Kosaraju, Mohammad Bavarian, Mark Chen, Heewoo Jun, Lukasz Kaiser, Matthias Plappert, Jerry Tworek, Jacob Hilton, Reiichiro Nakano, Christopher Hesse, and John Schulman. 2021.
\newblock \href {https://arxiv.org/abs/2110.14168} {Training verifiers to solve math word problems}.
\newblock \emph{Preprint}, arXiv:2110.14168.

\bibitem[{Das et~al.(2024)Das, Banerjee, Aditya, and Kulkarni}]{das-etal-2024-mathsensei}
Debrup Das, Debopriyo Banerjee, Somak Aditya, and Ashish Kulkarni. 2024.
\newblock \href {https://doi.org/10.18653/v1/2024.naacl-long.54} {{MATHSENSEI}: A tool-augmented large language model for mathematical reasoning}.
\newblock In \emph{Proceedings of the 2024 Conference of the North American Chapter of the Association for Computational Linguistics: Human Language Technologies (Volume 1: Long Papers)}, pages 942--966, Mexico City, Mexico. Association for Computational Linguistics.

\bibitem[{Ding et~al.(2023)Ding, Qin, Yang, Wei, Zonghan, Su, Hu, Chen, Chan, Chen, Yi, Zhao, Wang, Liu, Zheng, Chen, Liu, Tang, Li, and Sun}]{ding23naturemi}
Ning Ding, Yujia Qin, Guang Yang, Fuchao Wei, Yang Zonghan, Yusheng Su, Shengding Hu, Yulin Chen, Chi-Min Chan, Weize Chen, Jing Yi, Weilin Zhao, Xiaozhi Wang, Zhiyuan Liu, Hai-Tao Zheng, Jianfei Chen, Yang Liu, Jie Tang, Juanzi Li, and Maosong Sun. 2023.
\newblock \href {https://doi.org/10.1038/s42256-023-00626-4} {Parameter-efficient fine-tuning of large-scale pre-trained language models}.
\newblock \emph{Nature Machine Intelligence}, 5:1--16.

\bibitem[{Duarte et~al.(2024)Duarte, Zhao, Oliveira, and Li}]{duarte2024decop}
André~V. Duarte, Xuandong Zhao, Arlindo~L. Oliveira, and Lei Li. 2024.
\newblock \href {https://arxiv.org/abs/2402.09910} {{DE-COP}: Detecting copyrighted content in language models training data}.
\newblock \emph{Preprint}, arXiv:2402.09910.

\bibitem[{Dubey et~al.(2024)Dubey, Jauhri, Pandey, Kadian, Al-Dahle, Letman, Mathur, Schelten, Yang, Fan, Goyal, Hartshorn, Yang, Mitra, Sravankumar, Korenev, Hinsvark, Rao, Zhang, Rodriguez, Gregerson, Spataru, Roziere, Biron, Tang, Chern, Caucheteux, Nayak, Bi, Marra, McConnell, Keller, Touret, Wu, Wong, Ferrer, Nikolaidis, Allonsius, Song, Pintz, Livshits, Esiobu, Choudhary, Mahajan, Garcia-Olano, Perino, Hupkes, Lakomkin, AlBadawy, Lobanova, Dinan, Smith, Radenovic, Zhang, Synnaeve, Lee, Anderson, Nail, Mialon, Pang, Cucurell, Nguyen, Korevaar, Xu, Touvron, Zarov, Ibarra, Kloumann, Misra, Evtimov, Copet, Lee, Geffert, Vranes, Park, Mahadeokar, Shah, van~der Linde, Billock, Hong, Lee, Fu, Chi, Huang, Liu, Wang, Yu, Bitton, Spisak, Park, Rocca, Johnstun, Saxe, Jia, Alwala, Upasani, Plawiak, Li, Heafield, Stone, El-Arini, Iyer, Malik, Chiu, Bhalla, Rantala-Yeary, van~der Maaten, Chen, Tan, Jenkins, Martin, Madaan, Malo, Blecher, Landzaat, de~Oliveira, Muzzi, Pasupuleti, Singh, Paluri, Kardas, Oldham, Rita,
  Pavlova, Kambadur, Lewis, Si, Singh, Hassan, Goyal, Torabi, Bashlykov, Bogoychev, Chatterji, Duchenne, Çelebi, Alrassy, Zhang, Li, Vasic, Weng, Bhargava, Dubal, Krishnan, Koura, Xu, He, Dong, Srinivasan, Ganapathy, Calderer, Cabral, Stojnic, Raileanu, Girdhar, Patel, Sauvestre, Polidoro, Sumbaly, Taylor, Silva, Hou, Wang, Hosseini, Chennabasappa, Singh, Bell, Kim, Edunov, Nie, Narang, Raparthy, Shen, Wan, Bhosale, Zhang, Vandenhende, Batra, Whitman, Sootla, Collot, Gururangan, Borodinsky, Herman, Fowler, Sheasha, Georgiou, Scialom, Speckbacher, Mihaylov, Xiao, Karn, Goswami, Gupta, Ramanathan, Kerkez, Gonguet, Do, Vogeti, Petrovic, Chu, Xiong, Fu, Meers, Martinet, Wang, Tan, Xie, Jia, Wang, Goldschlag, Gaur, Babaei, Wen, Song, Zhang, Li, Mao, Coudert, Yan, Chen, Papakipos, Singh, Grattafiori, Jain, Kelsey, Shajnfeld, Gangidi, Victoria, Goldstand, Menon, Sharma, Boesenberg, Vaughan, Baevski, Feinstein, Kallet, Sangani, Yunus, Lupu, Alvarado, Caples, Gu, Ho, Poulton, Ryan, Ramchandani, Franco, Saraf,
  Chowdhury, Gabriel, Bharambe, Eisenman, Yazdan, James, Maurer, Leonhardi, Huang, Loyd, Paola, Paranjape, Liu, Wu, Ni, Hancock, Wasti, Spence, Stojkovic, Gamido, Montalvo, Parker, Burton, Mejia, Wang, Kim, Zhou, Hu, Chu, Cai, Tindal, Feichtenhofer, Civin, Beaty, Kreymer, Li, Wyatt, Adkins, Xu, Testuggine, David, Parikh, Liskovich, Foss, Wang, Le, Holland, Dowling, Jamil, Montgomery, Presani, Hahn, Wood, Brinkman, Arcaute, Dunbar, Smothers, Sun, Kreuk, Tian, Ozgenel, Caggioni, Guzmán, Kanayet, Seide, Florez, Schwarz, Badeer, Swee, Halpern, Thattai, Herman, Sizov, Guangyi, Zhang, Lakshminarayanan, Shojanazeri, Zou, Wang, Zha, Habeeb, Rudolph, Suk, Aspegren, Goldman, Damlaj, Molybog, Tufanov, Veliche, Gat, Weissman, Geboski, Kohli, Asher, Gaya, Marcus, Tang, Chan, Zhen, Reizenstein, Teboul, Zhong, Jin, Yang, Cummings, Carvill, Shepard, McPhie, Torres, Ginsburg, Wang, Wu, U, Saxena, Prasad, Khandelwal, Zand, Matosich, Veeraraghavan, Michelena, Li, Huang, Chawla, Lakhotia, Huang, Chen, Garg, A, Silva, Bell,
  Zhang, Guo, Yu, Moshkovich, Wehrstedt, Khabsa, Avalani, Bhatt, Tsimpoukelli, Mankus, Hasson, Lennie, Reso, Groshev, Naumov, Lathi, Keneally, Seltzer, Valko, Restrepo, Patel, Vyatskov, Samvelyan, Clark, Macey, Wang, Hermoso, Metanat, Rastegari, Bansal, Santhanam, Parks, White, Bawa, Singhal, Egebo, Usunier, Laptev, Dong, Zhang, Cheng, Chernoguz, Hart, Salpekar, Kalinli, Kent, Parekh, Saab, Balaji, Rittner, Bontrager, Roux, Dollar, Zvyagina, Ratanchandani, Yuvraj, Liang, Alao, Rodriguez, Ayub, Murthy, Nayani, Mitra, Li, Hogan, Battey, Wang, Maheswari, Howes, Rinott, Bondu, Datta, Chugh, Hunt, Dhillon, Sidorov, Pan, Verma, Yamamoto, Ramaswamy, Lindsay, Lindsay, Feng, Lin, Zha, Shankar, Zhang, Zhang, Wang, Agarwal, Sajuyigbe, Chintala, Max, Chen, Kehoe, Satterfield, Govindaprasad, Gupta, Cho, Virk, Subramanian, Choudhury, Goldman, Remez, Glaser, Best, Kohler, Robinson, Li, Zhang, Matthews, Chou, Shaked, Vontimitta, Ajayi, Montanez, Mohan, Kumar, Mangla, Albiero, Ionescu, Poenaru, Mihailescu, Ivanov, Li, Wang,
  Jiang, Bouaziz, Constable, Tang, Wang, Wu, Wang, Xia, Wu, Gao, Chen, Hu, Jia, Qi, Li, Zhang, Zhang, Adi, Nam, Yu, Wang, Hao, Qian, He, Rait, DeVito, Rosnbrick, Wen, Yang, and Zhao}]{dubey2024llama3herdmodels}
Abhimanyu Dubey, Abhinav Jauhri, Abhinav Pandey, Abhishek Kadian, Ahmad Al-Dahle, Aiesha Letman, Akhil Mathur, Alan Schelten, Amy Yang, Angela Fan, Anirudh Goyal, Anthony Hartshorn, Aobo Yang, Archi Mitra, Archie Sravankumar, Artem Korenev, Arthur Hinsvark, Arun Rao, Aston Zhang, Aurelien Rodriguez, Austen Gregerson, Ava Spataru, Baptiste Roziere, Bethany Biron, Binh Tang, Bobbie Chern, Charlotte Caucheteux, Chaya Nayak, Chloe Bi, Chris Marra, Chris McConnell, Christian Keller, Christophe Touret, Chunyang Wu, Corinne Wong, Cristian~Canton Ferrer, Cyrus Nikolaidis, Damien Allonsius, Daniel Song, Danielle Pintz, Danny Livshits, David Esiobu, Dhruv Choudhary, Dhruv Mahajan, Diego Garcia-Olano, Diego Perino, Dieuwke Hupkes, Egor Lakomkin, Ehab AlBadawy, Elina Lobanova, Emily Dinan, Eric~Michael Smith, Filip Radenovic, Frank Zhang, Gabriel Synnaeve, Gabrielle Lee, Georgia~Lewis Anderson, Graeme Nail, Gregoire Mialon, Guan Pang, Guillem Cucurell, Hailey Nguyen, Hannah Korevaar, Hu~Xu, Hugo Touvron, Iliyan Zarov,
  Imanol~Arrieta Ibarra, Isabel Kloumann, Ishan Misra, Ivan Evtimov, Jade Copet, Jaewon Lee, Jan Geffert, Jana Vranes, Jason Park, Jay Mahadeokar, Jeet Shah, Jelmer van~der Linde, Jennifer Billock, Jenny Hong, Jenya Lee, Jeremy Fu, Jianfeng Chi, Jianyu Huang, Jiawen Liu, Jie Wang, Jiecao Yu, Joanna Bitton, Joe Spisak, Jongsoo Park, Joseph Rocca, Joshua Johnstun, Joshua Saxe, Junteng Jia, Kalyan~Vasuden Alwala, Kartikeya Upasani, Kate Plawiak, Ke~Li, Kenneth Heafield, Kevin Stone, Khalid El-Arini, Krithika Iyer, Kshitiz Malik, Kuenley Chiu, Kunal Bhalla, Lauren Rantala-Yeary, Laurens van~der Maaten, Lawrence Chen, Liang Tan, Liz Jenkins, Louis Martin, Lovish Madaan, Lubo Malo, Lukas Blecher, Lukas Landzaat, Luke de~Oliveira, Madeline Muzzi, Mahesh Pasupuleti, Mannat Singh, Manohar Paluri, Marcin Kardas, Mathew Oldham, Mathieu Rita, Maya Pavlova, Melanie Kambadur, Mike Lewis, Min Si, Mitesh~Kumar Singh, Mona Hassan, Naman Goyal, Narjes Torabi, Nikolay Bashlykov, Nikolay Bogoychev, Niladri Chatterji, Olivier
  Duchenne, Onur Çelebi, Patrick Alrassy, Pengchuan Zhang, Pengwei Li, Petar Vasic, Peter Weng, Prajjwal Bhargava, Pratik Dubal, Praveen Krishnan, Punit~Singh Koura, Puxin Xu, Qing He, Qingxiao Dong, Ragavan Srinivasan, Raj Ganapathy, Ramon Calderer, Ricardo~Silveira Cabral, Robert Stojnic, Roberta Raileanu, Rohit Girdhar, Rohit Patel, Romain Sauvestre, Ronnie Polidoro, Roshan Sumbaly, Ross Taylor, Ruan Silva, Rui Hou, Rui Wang, Saghar Hosseini, Sahana Chennabasappa, Sanjay Singh, Sean Bell, Seohyun~Sonia Kim, Sergey Edunov, Shaoliang Nie, Sharan Narang, Sharath Raparthy, Sheng Shen, Shengye Wan, Shruti Bhosale, Shun Zhang, Simon Vandenhende, Soumya Batra, Spencer Whitman, Sten Sootla, Stephane Collot, Suchin Gururangan, Sydney Borodinsky, Tamar Herman, Tara Fowler, Tarek Sheasha, Thomas Georgiou, Thomas Scialom, Tobias Speckbacher, Todor Mihaylov, Tong Xiao, Ujjwal Karn, Vedanuj Goswami, Vibhor Gupta, Vignesh Ramanathan, Viktor Kerkez, Vincent Gonguet, Virginie Do, Vish Vogeti, Vladan Petrovic, Weiwei Chu,
  Wenhan Xiong, Wenyin Fu, Whitney Meers, Xavier Martinet, Xiaodong Wang, Xiaoqing~Ellen Tan, Xinfeng Xie, Xuchao Jia, Xuewei Wang, Yaelle Goldschlag, Yashesh Gaur, Yasmine Babaei, Yi~Wen, Yiwen Song, Yuchen Zhang, Yue Li, Yuning Mao, Zacharie~Delpierre Coudert, Zheng Yan, Zhengxing Chen, Zoe Papakipos, Aaditya Singh, Aaron Grattafiori, Abha Jain, Adam Kelsey, Adam Shajnfeld, Adithya Gangidi, Adolfo Victoria, Ahuva Goldstand, Ajay Menon, Ajay Sharma, Alex Boesenberg, Alex Vaughan, Alexei Baevski, Allie Feinstein, Amanda Kallet, Amit Sangani, Anam Yunus, Andrei Lupu, Andres Alvarado, Andrew Caples, Andrew Gu, Andrew Ho, Andrew Poulton, Andrew Ryan, Ankit Ramchandani, Annie Franco, Aparajita Saraf, Arkabandhu Chowdhury, Ashley Gabriel, Ashwin Bharambe, Assaf Eisenman, Azadeh Yazdan, Beau James, Ben Maurer, Benjamin Leonhardi, Bernie Huang, Beth Loyd, Beto~De Paola, Bhargavi Paranjape, Bing Liu, Bo~Wu, Boyu Ni, Braden Hancock, Bram Wasti, Brandon Spence, Brani Stojkovic, Brian Gamido, Britt Montalvo, Carl
  Parker, Carly Burton, Catalina Mejia, Changhan Wang, Changkyu Kim, Chao Zhou, Chester Hu, Ching-Hsiang Chu, Chris Cai, Chris Tindal, Christoph Feichtenhofer, Damon Civin, Dana Beaty, Daniel Kreymer, Daniel Li, Danny Wyatt, David Adkins, David Xu, Davide Testuggine, Delia David, Devi Parikh, Diana Liskovich, Didem Foss, Dingkang Wang, Duc Le, Dustin Holland, Edward Dowling, Eissa Jamil, Elaine Montgomery, Eleonora Presani, Emily Hahn, Emily Wood, Erik Brinkman, Esteban Arcaute, Evan Dunbar, Evan Smothers, Fei Sun, Felix Kreuk, Feng Tian, Firat Ozgenel, Francesco Caggioni, Francisco Guzmán, Frank Kanayet, Frank Seide, Gabriela~Medina Florez, Gabriella Schwarz, Gada Badeer, Georgia Swee, Gil Halpern, Govind Thattai, Grant Herman, Grigory Sizov, Guangyi, Zhang, Guna Lakshminarayanan, Hamid Shojanazeri, Han Zou, Hannah Wang, Hanwen Zha, Haroun Habeeb, Harrison Rudolph, Helen Suk, Henry Aspegren, Hunter Goldman, Ibrahim Damlaj, Igor Molybog, Igor Tufanov, Irina-Elena Veliche, Itai Gat, Jake Weissman, James
  Geboski, James Kohli, Japhet Asher, Jean-Baptiste Gaya, Jeff Marcus, Jeff Tang, Jennifer Chan, Jenny Zhen, Jeremy Reizenstein, Jeremy Teboul, Jessica Zhong, Jian Jin, Jingyi Yang, Joe Cummings, Jon Carvill, Jon Shepard, Jonathan McPhie, Jonathan Torres, Josh Ginsburg, Junjie Wang, Kai Wu, Kam~Hou U, Karan Saxena, Karthik Prasad, Kartikay Khandelwal, Katayoun Zand, Kathy Matosich, Kaushik Veeraraghavan, Kelly Michelena, Keqian Li, Kun Huang, Kunal Chawla, Kushal Lakhotia, Kyle Huang, Lailin Chen, Lakshya Garg, Lavender A, Leandro Silva, Lee Bell, Lei Zhang, Liangpeng Guo, Licheng Yu, Liron Moshkovich, Luca Wehrstedt, Madian Khabsa, Manav Avalani, Manish Bhatt, Maria Tsimpoukelli, Martynas Mankus, Matan Hasson, Matthew Lennie, Matthias Reso, Maxim Groshev, Maxim Naumov, Maya Lathi, Meghan Keneally, Michael~L. Seltzer, Michal Valko, Michelle Restrepo, Mihir Patel, Mik Vyatskov, Mikayel Samvelyan, Mike Clark, Mike Macey, Mike Wang, Miquel~Jubert Hermoso, Mo~Metanat, Mohammad Rastegari, Munish Bansal, Nandhini
  Santhanam, Natascha Parks, Natasha White, Navyata Bawa, Nayan Singhal, Nick Egebo, Nicolas Usunier, Nikolay~Pavlovich Laptev, Ning Dong, Ning Zhang, Norman Cheng, Oleg Chernoguz, Olivia Hart, Omkar Salpekar, Ozlem Kalinli, Parkin Kent, Parth Parekh, Paul Saab, Pavan Balaji, Pedro Rittner, Philip Bontrager, Pierre Roux, Piotr Dollar, Polina Zvyagina, Prashant Ratanchandani, Pritish Yuvraj, Qian Liang, Rachad Alao, Rachel Rodriguez, Rafi Ayub, Raghotham Murthy, Raghu Nayani, Rahul Mitra, Raymond Li, Rebekkah Hogan, Robin Battey, Rocky Wang, Rohan Maheswari, Russ Howes, Ruty Rinott, Sai~Jayesh Bondu, Samyak Datta, Sara Chugh, Sara Hunt, Sargun Dhillon, Sasha Sidorov, Satadru Pan, Saurabh Verma, Seiji Yamamoto, Sharadh Ramaswamy, Shaun Lindsay, Shaun Lindsay, Sheng Feng, Shenghao Lin, Shengxin~Cindy Zha, Shiva Shankar, Shuqiang Zhang, Shuqiang Zhang, Sinong Wang, Sneha Agarwal, Soji Sajuyigbe, Soumith Chintala, Stephanie Max, Stephen Chen, Steve Kehoe, Steve Satterfield, Sudarshan Govindaprasad, Sumit Gupta,
  Sungmin Cho, Sunny Virk, Suraj Subramanian, Sy~Choudhury, Sydney Goldman, Tal Remez, Tamar Glaser, Tamara Best, Thilo Kohler, Thomas Robinson, Tianhe Li, Tianjun Zhang, Tim Matthews, Timothy Chou, Tzook Shaked, Varun Vontimitta, Victoria Ajayi, Victoria Montanez, Vijai Mohan, Vinay~Satish Kumar, Vishal Mangla, Vítor Albiero, Vlad Ionescu, Vlad Poenaru, Vlad~Tiberiu Mihailescu, Vladimir Ivanov, Wei Li, Wenchen Wang, Wenwen Jiang, Wes Bouaziz, Will Constable, Xiaocheng Tang, Xiaofang Wang, Xiaojian Wu, Xiaolan Wang, Xide Xia, Xilun Wu, Xinbo Gao, Yanjun Chen, Ye~Hu, Ye~Jia, Ye~Qi, Yenda Li, Yilin Zhang, Ying Zhang, Yossi Adi, Youngjin Nam, Yu, Wang, Yuchen Hao, Yundi Qian, Yuzi He, Zach Rait, Zachary DeVito, Zef Rosnbrick, Zhaoduo Wen, Zhenyu Yang, and Zhiwei Zhao. 2024.
\newblock \href {https://arxiv.org/abs/2407.21783} {The {L}lama 3 herd of models}.
\newblock \emph{Preprint}, arXiv:2407.21783.

\bibitem[{Gao et~al.(2023)Gao, Madaan, Zhou, Alon, Liu, Yang, Callan, and Neubig}]{10.5555/3618408.3618843}
Luyu Gao, Aman Madaan, Shuyan Zhou, Uri Alon, Pengfei Liu, Yiming Yang, Jamie Callan, and Graham Neubig. 2023.
\newblock \href {https://dl.acm.org/doi/10.5555/3618408.3618843} {{PAL}: Program-aided language models}.
\newblock In \emph{Proceedings of the 40th International Conference on Machine Learning}, ICML'23. JMLR.org.

\bibitem[{Gou et~al.(2024)Gou, Shao, Gong, Shen, Yang, Huang, Duan, and Chen}]{gou2024toratoolintegratedreasoningagent}
Zhibin Gou, Zhihong Shao, Yeyun Gong, Yelong Shen, Yujiu Yang, Minlie Huang, Nan Duan, and Weizhu Chen. 2024.
\newblock \href {https://arxiv.org/abs/2309.17452} {To{RA}: A tool-integrated reasoning agent for mathematical problem solving}.
\newblock In \emph{Proceedings of the Twelfth International Conference on Learning Representations}, ICLR'24.

\bibitem[{Goyal et~al.(2024)Goyal, Ji, Rawat, Menon, Kumar, and Nagarajan}]{goyal2024thinkspeaktraininglanguage}
Sachin Goyal, Ziwei Ji, Ankit~Singh Rawat, Aditya~Krishna Menon, Sanjiv Kumar, and Vaishnavh Nagarajan. 2024.
\newblock \href {https://arxiv.org/abs/2310.02226} {Think before you speak: Training language models with pause tokens}.
\newblock In \emph{Proceedings of the Twelfth International Conference on Learning Representations}, ICLR'24.

\bibitem[{Gupta and Kembhavi(2023)}]{10204174}
Tanmay Gupta and Aniruddha Kembhavi. 2023.
\newblock \href {https://doi.org/10.1109/CVPR52729.2023.01436} {Visual programming: Compositional visual reasoning without training}.
\newblock In \emph{2023 IEEE/CVF Conference on Computer Vision and Pattern Recognition (CVPR)}, pages 14953--14962.

\bibitem[{Han et~al.(2024)Han, Gao, Liu, Zhang, and Zhang}]{han2024parameterefficient}
Zeyu Han, Chao Gao, Jinyang Liu, Jeff Zhang, and Sai~Qian Zhang. 2024.
\newblock \href {https://arxiv.org/abs/2403.14608} {Parameter-efficient fine-tuning for large models: A comprehensive survey}.
\newblock \emph{Preprint}, arXiv:2403.14608.

\bibitem[{Hao et~al.(2023)Hao, Liu, Wang, and Hu}]{NEURIPS2023_8fd1a81c}
Shibo Hao, Tianyang Liu, Zhen Wang, and Zhiting Hu. 2023.
\newblock \href {https://proceedings.neurips.cc/paper_files/paper/2023/file/8fd1a81c882cd45f64958da6284f4a3f-Paper-Conference.pdf} {Toolken{GPT}: Augmenting frozen language models with massive tools via tool embeddings}.
\newblock In \emph{Advances in Neural Information Processing Systems}, volume~36, pages 45870--45894. Curran Associates, Inc.

\bibitem[{Herel and Mikolov(2024)}]{herel2024thinkingtokenslanguagemodeling}
David Herel and Tomas Mikolov. 2024.
\newblock \href {https://arxiv.org/abs/2405.08644} {Thinking tokens for language modeling}.
\newblock \emph{Preprint}, arXiv:2405.08644.

\bibitem[{Hsieh et~al.(2023)Hsieh, Chen, Li, Fujii, Ratner, Lee, Krishna, and Pfister}]{hsieh2023tool}
Cheng-Yu Hsieh, Si-An Chen, Chun-Liang Li, Yasuhisa Fujii, Alexander Ratner, Chen-Yu Lee, Ranjay Krishna, and Tomas Pfister. 2023.
\newblock \href {https://arxiv.org/abs/2308.00675} {Tool documentation enables zero-shot tool-usage with large language models}.
\newblock \emph{Preprint}, arXiv:2308.00675.

\bibitem[{Kojima et~al.(2022)Kojima, Gu, Reid, Matsuo, and Iwasawa}]{NEURIPS2022_8bb0d291}
Takeshi Kojima, Shixiang~(Shane) Gu, Machel Reid, Yutaka Matsuo, and Yusuke Iwasawa. 2022.
\newblock \href {https://proceedings.neurips.cc/paper_files/paper/2022/file/8bb0d291acd4acf06ef112099c16f326-Paper-Conference.pdf} {Large language models are zero-shot reasoners}.
\newblock In \emph{Advances in Neural Information Processing Systems}, volume~35, pages 22199--22213. Curran Associates, Inc.

\bibitem[{Koncel-Kedziorski et~al.(2016)Koncel-Kedziorski, Roy, Amini, Kushman, and Hajishirzi}]{koncel-kedziorski-etal-2016-mawps}
Rik Koncel-Kedziorski, Subhro Roy, Aida Amini, Nate Kushman, and Hannaneh Hajishirzi. 2016.
\newblock \href {https://doi.org/10.18653/v1/N16-1136} {{MAWPS}: A math word problem repository}.
\newblock In \emph{Proceedings of the 2016 Conference of the North {A}merican Chapter of the Association for Computational Linguistics: Human Language Technologies}, pages 1152--1157, San Diego, California. Association for Computational Linguistics.

\bibitem[{Kudo and Richardson(2018)}]{kudo-richardson-2018-sentencepiece}
Taku Kudo and John Richardson. 2018.
\newblock \href {https://doi.org/10.18653/v1/D18-2012} {{S}entence{P}iece: A simple and language independent subword tokenizer and detokenizer for neural text processing}.
\newblock In \emph{Proceedings of the 2018 Conference on Empirical Methods in Natural Language Processing: System Demonstrations}, pages 66--71, Brussels, Belgium. Association for Computational Linguistics.

\bibitem[{Miao et~al.(2020)Miao, Liang, and Su}]{miao-etal-2020-diverse}
Shen-yun Miao, Chao-Chun Liang, and Keh-Yih Su. 2020.
\newblock \href {https://doi.org/10.18653/v1/2020.acl-main.92} {A diverse corpus for evaluating and developing {E}nglish math word problem solvers}.
\newblock In \emph{Proceedings of the 58th Annual Meeting of the Association for Computational Linguistics}, pages 975--984, Online. Association for Computational Linguistics.

\bibitem[{Mishra et~al.(2022)Mishra, Finlayson, Lu, Tang, Welleck, Baral, Rajpurohit, Tafjord, Sabharwal, Clark, and Kalyan}]{Mishra2022Lila}
Swaroop Mishra, Matthew Finlayson, Pan Lu, Leonard Tang, Sean Welleck, Chitta Baral, Tanmay Rajpurohit, Oyvind Tafjord, Ashish Sabharwal, Peter Clark, and Ashwin Kalyan. 2022.
\newblock \href {https://aclanthology.org/2022.emnlp-main.392/} {Lila: A unified benchmark for mathematical reasoning}.
\newblock In \emph{Proceedings of the 2022 Conference on Empirical Methods in Natural Language Processing (EMNLP)}.

\bibitem[{Ott et~al.(2023)Ott, Hebenstreit, Liévin, Egeberg~Hother, Moradi, Mayrhauser, Praas, Winther, and Samwald}]{hub_nature}
Simon Ott, Konstantin Hebenstreit, Valentin Liévin, Christoffer Egeberg~Hother, Milad Moradi, Maximilian Mayrhauser, Robert Praas, Ole Winther, and Matthias Samwald. 2023.
\newblock \href {https://doi.org/10.1038/s41597-023-02433-3} {Thoughtsource: A central hub for large language model reasoning data}.
\newblock \emph{Scientific Data}, 10:528.

\bibitem[{Paranjape et~al.(2023)Paranjape, Lundberg, Singh, Hajishirzi, Zettlemoyer, and Ribeiro}]{paranjape2023art}
Bhargavi Paranjape, Scott Lundberg, Sameer Singh, Hannaneh Hajishirzi, Luke Zettlemoyer, and Marco~Tulio Ribeiro. 2023.
\newblock \href {https://arxiv.org/abs/2303.09014} {{ART}: Automatic multi-step reasoning and tool-use for large language models}.
\newblock \emph{Preprint}, arXiv:2303.09014.

\bibitem[{Parisi et~al.(2022)Parisi, Zhao, and Fiedel}]{parisi2022talm}
Aaron Parisi, Yao Zhao, and Noah Fiedel. 2022.
\newblock \href {https://arxiv.org/abs/2205.12255} {{TALM}: Tool augmented language models}.
\newblock \emph{Preprint}, arXiv:2205.12255.

\bibitem[{Patel et~al.(2021)Patel, Bhattamishra, and Goyal}]{patel-etal-2021-nlp}
Arkil Patel, Satwik Bhattamishra, and Navin Goyal. 2021.
\newblock \href {https://doi.org/10.18653/v1/2021.naacl-main.168} {Are {NLP} models really able to solve simple math word problems?}
\newblock In \emph{Proceedings of the 2021 Conference of the North American Chapter of the Association for Computational Linguistics: Human Language Technologies}, pages 2080--2094, Online. Association for Computational Linguistics.

\bibitem[{Patil et~al.(2023)Patil, Zhang, Wang, and Gonzalez}]{patil2023gorilla}
Shishir~G. Patil, Tianjun Zhang, Xin Wang, and Joseph~E. Gonzalez. 2023.
\newblock \href {https://arxiv.org/abs/2305.15334} {Gorilla: Large language model connected with massive {API}s}.
\newblock \emph{Preprint}, arXiv:2305.15334.

\bibitem[{Qiao et~al.(2024)Qiao, Gui, Jia, Chen, and Zhang}]{qiao2024trice}
Shuofei Qiao, Honghao Gui, Qianghuai Jia, Huajun Chen, and Ningyu Zhang. 2024.
\newblock \href {https://aclanthology.org/2024.naacl-long.195/} {Making language models better tool learners with execution feedback}.
\newblock In \emph{Annual Conference of the North American Chapter of the Association for Computational Linguistics (NAACL)}.

\bibitem[{Schick et~al.(2023)Schick, Dwivedi-Yu, Dessi, Raileanu, Lomeli, Hambro, Zettlemoyer, Cancedda, and Scialom}]{NEURIPS2023_d842425e}
Timo Schick, Jane Dwivedi-Yu, Roberto Dessi, Roberta Raileanu, Maria Lomeli, Eric Hambro, Luke Zettlemoyer, Nicola Cancedda, and Thomas Scialom. 2023.
\newblock \href {https://proceedings.neurips.cc/paper_files/paper/2023/file/d842425e4bf79ba039352da0f658a906-Paper-Conference.pdf} {Toolformer: Language models can teach themselves to use tools}.
\newblock In \emph{Advances in Neural Information Processing Systems}, volume~36, pages 68539--68551. Curran Associates, Inc.

\bibitem[{Sennrich et~al.(2016)Sennrich, Haddow, and Birch}]{sennrich-etal-2016-neural}
Rico Sennrich, Barry Haddow, and Alexandra Birch. 2016.
\newblock \href {https://doi.org/10.18653/v1/P16-1162} {Neural machine translation of rare words with subword units}.
\newblock In \emph{Proceedings of the 54th Annual Meeting of the Association for Computational Linguistics (Volume 1: Long Papers)}, pages 1715--1725, Berlin, Germany. Association for Computational Linguistics.

\bibitem[{Shinn et~al.(2023)Shinn, Cassano, Gopinath, Narasimhan, and Yao}]{NEURIPS2023_1b44b878}
Noah Shinn, Federico Cassano, Ashwin Gopinath, Karthik Narasimhan, and Shunyu Yao. 2023.
\newblock \href {https://proceedings.neurips.cc/paper_files/paper/2023/file/1b44b878bb782e6954cd888628510e90-Paper-Conference.pdf} {Reflexion: language agents with verbal reinforcement learning}.
\newblock In \emph{Advances in Neural Information Processing Systems}, volume~36, pages 8634--8652. Curran Associates, Inc.

\bibitem[{Sur\'is et~al.(2023)Sur\'is, Menon, and Vondrick}]{surismenon2023vipergpt}
D\'idac Sur\'is, Sachit Menon, and Carl Vondrick. 2023.
\newblock \href {https://openaccess.thecvf.com/content/ICCV2023/papers/Suris_ViperGPT_Visual_Inference_via_Python_Execution_for_Reasoning_ICCV_2023_paper.pdf} {Viper{GPT}: Visual inference via python execution for reasoning}.
\newblock \emph{Proceedings of IEEE International Conference on Computer Vision (ICCV)}.

\bibitem[{Tang et~al.(2023)Tang, Deng, Lin, Han, Liang, and Sun}]{tang2023toolalpaca}
Qiaoyu Tang, Ziliang Deng, Hongyu Lin, Xianpei Han, Qiao Liang, and Le~Sun. 2023.
\newblock \href {https://arxiv.org/abs/2306.05301} {Tool{A}lpaca: Generalized tool learning for language models with 3000 simulated cases}.
\newblock \emph{Preprint}, arXiv:2306.05301.

\bibitem[{Taori et~al.(2023)Taori, Gulrajani, Zhang, Dubois, Li, Guestrin, Liang, and Hashimoto}]{alpacareport}
Rohan Taori, Ishaan Gulrajani, Tianyi Zhang, Yann Dubois, Xuechen Li, Carlos Guestrin, Percy Liang, and Tatsunori~B Hashimoto. 2023.
\newblock \href {https://crfm.stanford.edu/2023/03/13/alpaca.html} {Alpaca: A strong, replicable instruction-following model}.

\bibitem[{Touvron et~al.(2023)Touvron, Lavril, Izacard, Martinet, Lachaux, Lacroix, Rozière, Goyal, Hambro, Azhar, Rodriguez, Joulin, Grave, and Lample}]{touvron2023llama}
Hugo Touvron, Thibaut Lavril, Gautier Izacard, Xavier Martinet, Marie-Anne Lachaux, Timothée Lacroix, Baptiste Rozière, Naman Goyal, Eric Hambro, Faisal Azhar, Aurelien Rodriguez, Armand Joulin, Edouard Grave, and Guillaume Lample. 2023.
\newblock \href {https://arxiv.org/abs/2302.13971} {{LL}a{M}a: Open and efficient foundation language models}.
\newblock \emph{Preprint}, arXiv:2302.13971.

\bibitem[{Wang et~al.(2024)Wang, Caccia, Ostapenko, Yuan, and Sordoni}]{wang2024guiding}
Xinyi Wang, Lucas Caccia, O.~Ostapenko, Xingdi Yuan, and Alessandro Sordoni. 2024.
\newblock \href {https://arxiv.org/abs/2305.15334} {Guiding language model reasoning with planning tokens}.
\newblock \emph{Preprint}, arXiv:2305.15334.

\bibitem[{Wei et~al.(2022)Wei, Wang, Schuurmans, Bosma, ichter, Xia, Chi, Le, and Zhou}]{NEURIPS2022_9d560961}
Jason Wei, Xuezhi Wang, Dale Schuurmans, Maarten Bosma, brian ichter, Fei Xia, Ed~Chi, Quoc~V Le, and Denny Zhou. 2022.
\newblock \href {https://proceedings.neurips.cc/paper_files/paper/2022/file/9d5609613524ecf4f15af0f7b31abca4-Paper-Conference.pdf} {Chain-of-thought prompting elicits reasoning in large language models}.
\newblock In \emph{Advances in Neural Information Processing Systems}, volume~35, pages 24824--24837. Curran Associates, Inc.

\bibitem[{Yao et~al.(2023{\natexlab{a}})Yao, Yu, Zhao, Shafran, Griffiths, Cao, and Narasimhan}]{NEURIPS2023_271db992}
Shunyu Yao, Dian Yu, Jeffrey Zhao, Izhak Shafran, Tom Griffiths, Yuan Cao, and Karthik Narasimhan. 2023{\natexlab{a}}.
\newblock \href {https://proceedings.neurips.cc/paper_files/paper/2023/file/271db9922b8d1f4dd7aaef84ed5ac703-Paper-Conference.pdf} {Tree of {T}houghts: Deliberate problem solving with large language models}.
\newblock In \emph{Advances in Neural Information Processing Systems}, volume~36, pages 11809--11822. Curran Associates, Inc.

\bibitem[{Yao et~al.(2023{\natexlab{b}})Yao, Zhao, Yu, Du, Shafran, Narasimhan, and Cao}]{react}
Shunyu Yao, Jeffrey Zhao, Dian Yu, Nan Du, Izhak Shafran, Karthik Narasimhan, and Yuan Cao. 2023{\natexlab{b}}.
\newblock \href {https://arxiv.org/pdf/2210.03629} {Re{A}ct: Synergizing reasoning and acting in language models}.
\newblock In \emph{Proceedings of the Eleventh International Conference on Learning Representations}, ICLR'23.

\bibitem[{Zeng et~al.(2024{\natexlab{a}})Zeng, Xu, Wang, Zhang, Yin, Zhang, Rojas, Feng, Zhao, Lai, Yu, Wang, Sun, Zhang, Cheng, Gui, Tang, Zhang, Sun, Li, Zhao, Wu, Zhong, Liu, Huang, Zhang, Zheng, Lu, Duan, Zhang, Cao, Yang, Tam, Zhao, Liu, Xia, Zhang, Gu, Lv, Liu, Liu, Yang, Song, Zhang, An, Xu, Niu, Yang, Li, Bai, Dong, Qi, Wang, Yang, Du, Hou, and Wang}]{glm2024chatglmfamilylargelanguage}
Aohan Zeng, Bin Xu, Bowen Wang, Chenhui Zhang, Da~Yin, Dan Zhang, Diego Rojas, Guanyu Feng, Hanlin Zhao, Hanyu Lai, Hao Yu, Hongning Wang, Jiadai Sun, Jiajie Zhang, Jiale Cheng, Jiayi Gui, Jie Tang, Jing Zhang, Jingyu Sun, Juanzi Li, Lei Zhao, Lindong Wu, Lucen Zhong, Mingdao Liu, Minlie Huang, Peng Zhang, Qinkai Zheng, Rui Lu, Shuaiqi Duan, Shudan Zhang, Shulin Cao, Shuxun Yang, Weng~Lam Tam, Wenyi Zhao, Xiao Liu, Xiao Xia, Xiaohan Zhang, Xiaotao Gu, Xin Lv, Xinghan Liu, Xinyi Liu, Xinyue Yang, Xixuan Song, Xunkai Zhang, Yifan An, Yifan Xu, Yilin Niu, Yuantao Yang, Yueyan Li, Yushi Bai, Yuxiao Dong, Zehan Qi, Zhaoyu Wang, Zhen Yang, Zhengxiao Du, Zhenyu Hou, and Zihan Wang. 2024{\natexlab{a}}.
\newblock \href {https://arxiv.org/abs/2406.12793} {Chat{GLM}: A family of large language models from {GLM}-130b to {GLM}-4 {A}ll {T}ools}.
\newblock \emph{Preprint}, arXiv:2406.12793.

\bibitem[{Zeng et~al.(2024{\natexlab{b}})Zeng, Chen, Liu, Jiang, and Jia}]{zeng2024mrgsm8kmetareasoningbenchmarklarge}
Zhongshen Zeng, Pengguang Chen, Shu Liu, Haiyun Jiang, and Jiaya Jia. 2024{\natexlab{b}}.
\newblock \href {https://arxiv.org/abs/2312.17080} {{MR}-{GSM8K}: A meta-reasoning benchmark for large language model evaluation}.
\newblock \emph{Preprint}, arXiv:2312.17080.

\bibitem[{Zheng et~al.(2024)Zheng, Zhou, Meng, Zhou, and Huang}]{llm-mcq-bias}
Chujie Zheng, Hao Zhou, Fandong Meng, Jie Zhou, and Minlie Huang. 2024.
\newblock \href {https://openreview.net/forum?id=shr9PXz7T0} {Large language models are not robust multiple choice selectors}.
\newblock In \emph{The Twelfth International Conference on Learning Representations}.

\bibitem[{Zheng et~al.(2023)Zheng, Zhang, Chiang, Li, Lin, Sheng, Wu, Zhuang, Zhuang, E, Stoica, and Xing}]{vicunareport}
Lianmin Zheng, Hao Zhang, Wei-Lin Chiang, Zhuohan Li, Zi~Lin, Ying Sheng, Zhanghao Wu, Siyuan Zhuang, Yonghao Zhuang, Gonzalez~Joseph E, Ion Stoica, and Eric~P Xing. 2023.
\newblock \href {https://lmsys.org/blog/2023-03-30-vicuna} {Vicuna: An open-source chatbot impressing {GPT}-4 with 90\%* {C}hat{GPT} quality}.

\end{thebibliography}

\clearpage

\appendix

\section{Preliminary Experiments}\label{sec:sampling}

\subsection{Top-\textit{k} Decoding} 
\begin{figure}[b!]
    \includegraphics[width=\linewidth]{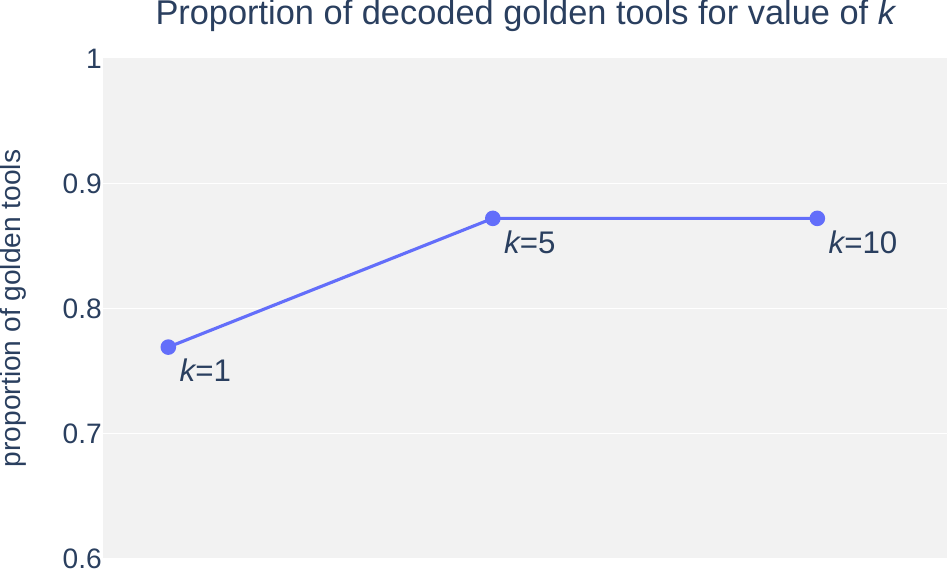}
    \caption{Proportion of golden tools across the FuncQA validation set, for \textit{k}=1, \textit{k}=5 and \textit{k}=10, where \textit{k}  is the number of top-probability tokens at each decoding step.}
    \label{fig:lineplot}
\end{figure}

We run ToolkenGPT (in its original implementation based on the first version of LLaMA) on the validation set of FuncQA. We find that in samples where the system has generated an incorrect tool by greedy sampling, the correct one---complete with correct arguments---is among the top five most likely tokens in over $60\%$ of cases. These include samples where the correct tool can be found by searching the top tokens at a \textit{different} position from the one that has produced the incorrect tool. Overall, the system decodes the correct tool (regardless of the arguments it generates for it) in 76.9\% of samples; this rises by over $10\%$ to $87.2\%$ when we consider the top $k=5$ tokens at each decoding step (Fig. \ref{fig:lineplot}). Further increasing $k$ to $10$ incurs a higher computational cost without raising the proportion of decoded golden tools.

\subsection{Temperature Sampling} We experiment with temperature sampling and find that it is not an optimal strategy to gather candidate tools, as lower temperatures do not lead to enough diversity while higher ones generate irrelevant tools.

\subsection{Tool Selection by Self-Consistency}

We select tools by self-consistency on FuncQA's validation set. We try choosing the tool that is most represented in the candidate set both before and after argument insertion. In both cases, We find that tool choice by self-consistency performs poorly (over 6\% below ToolkenGPT in the best case).

\section{Model Implementation and Training}
\label{sec:model}

We do all training and inference on a single NVIDIA Tesla V100 GPU.

\subsection{Tools}

GSM8K-XL and FuncQA \cite{NEURIPS2023_8fd1a81c} are annotated with four and 13 tools respectively, each representing a math operation. We illustrate these in Table~\ref{tab:tools}. Tools are trained as additional tokens added to the standard language modelling head of Llama 3, which comprises 128,256 token representations. Therefore, we extend these representations to 128,269 in the case of FuncQA and 128,260 for GSM8K-XL.

\subsection{Details of the Training Process}

The LM head of \textsc{Tecton} consists of the standard head of Llama 3 8B Instruct \cite{dubey2024llama3herdmodels} concatenated with an additional linear layer of size $embedding\_dimension \times number\_of\_tools$. The tool token embeddings are randomly initialized and trained on math reasoning QA pairs from GSM8K-XL and FuncQA, following a standard language modelling objective. Both datasets are annotated with the token positions at which each tool should be generated. Note that the original datasets made available by \citet{NEURIPS2023_8fd1a81c} are annotated according to SentencePiece tokenization \cite{kudo-richardson-2018-sentencepiece}. We edit the annotations to be compatible with Llama 3's Byte-Pair Encoding tokenizer \cite{sennrich-etal-2016-neural}. We tune two distinct sets of special tokens, one on each dataset, as each requires a different set of tools as shown in Table~\ref{tab:tools}. We train at half precision (FP16) for ten epochs, using learning rates \{$1$e$-4$, $1$e$-3$\}, batch size 1, and saving checkpoints at each epoch. We select the best checkpoint by measuring performance on a validation set.

Table \ref{tab:trainingdata} gives an overview of our training, validation, and testing data. It should be noted that GSM8K-XL's training and validation sets coincide with those of GSM8K, as only the test portion of the dataset was enhanced by \citet{NEURIPS2023_8fd1a81c}. Table \ref{tab:checkpoints} reports the hyperparameter combinations of our chosen checkpoints.

\begin{table}[b!]
\centering
\renewcommand\arraystretch{1.5}
\scalebox{0.9}{
\begin{tabular}{ll}
\hline
\toprule
\textbf{Task} & \textbf{Tools} \\
\hline
\textbf{\makecell[l]{GSM8K-XL}} & \makecell[l]{ \texttt{<add>} \\ \texttt{<subtract>} \\ \texttt{<multiply>} \\ \texttt{<divide>}} \\

\hline

\textbf{\makecell[l]{FuncQA}} & \makecell[l]{ \texttt{<add>} \\ \texttt{<subtract>} \\ \texttt{<multiply>} \\ \texttt{<divide>} \\
\texttt{<power>} \\ \texttt{<sqrt>} \\ \texttt{<log>} \\ \texttt{<ln>} \\
\texttt{<lcm>} \\ \texttt{<gcd>} \\ \texttt{<remainder>} \\ \texttt{<choose>} \\ \texttt{<permutate>} } \\

\bottomrule
\hline
\end{tabular}
}
\caption{Math reasoning datasets and their corresponding tools.}
\label{tab:tools}
\vspace{12pt}
\end{table}

\begin{table}[b!]
\centering
\renewcommand\arraystretch{2}
\scalebox{0.80}{
\begin{tabular}{lccc}
\hline
\toprule
\textbf{Dataset} & \textbf{Train} & \textbf{Validation} & \textbf{Test} \\
\hline
\textbf{GSM8K-XL} & {5054} & {1000} & {568} \\
\textbf{FuncQA} & {611} & {39} & {128} \\
\textbf{ASDiv-XL} & {N/A} & {N/A} & {360} \\
\textbf{MAWPS-XL} & {N/A} & {N/A} & {416} \\
\textbf{SVAMP-XL} & {N/A} & {N/A} & {270} \\
\bottomrule
\hline
\end{tabular}
}
\caption{Sizes of our training and testing datasets. Note that we use ASDiv-XL, MAWPS-XL and SVAMP-XL only for testing. For FuncQA the reported test data size includes both splits---FuncQA-OH and FuncQA-MH.}
\label{tab:trainingdata}
\end{table}

\begin{table}[b!]
\vspace{15pt}
\centering
\renewcommand\arraystretch{1.5}
\scalebox{0.80}{
\begin{tabular}{lccc}
\hline
\toprule
\textbf{Dataset} & \textbf{Epoch} & \textbf{Learning Rate} \\
\hline
\textbf{GSM8K-XL} & {9} & {1e$-3$} & \\
\textbf{FuncQA} & {4} & {1e$-4$} & \\
\bottomrule
\hline
\end{tabular}
}
\caption{Hyperparameter combinations of the chosen model checkpoints for GSM8K-XL and FuncQA.}
\label{tab:checkpoints}
\end{table}

\subsection{Inference-time Hyperparameters}

At inference, we decode with temperature~$t=0$, $p = 0.95$, $k = 5$. In the reasoning phase, we apply logit bias to the tool tokens to promote their generation. We use logit bias $3.0$ for GSM8K-XL and $4.0$ for FuncQA. Since our generation strategy is deterministic, all our accuracies are reported on a single run, rounded to 1 decimal place.

\section{Baselines Implementation}
\label{sec:baselines}

All baselines are built upon Llama 3 8B Instruct.

\paragraph{Llama 3 8B Instruct.} We measure our base model's performance in the zero-shot setting. We experiment with CoT zero-shot prompting (prepending to the question the instruction \texttt{Let's think step by step}) but find that just using the raw question as input results in better performance. 

\paragraph{Chain-of-Thought (CoT).} We extract from the training set six pairs of questions and answers demonstrating maths operations and their use. For comparability, we use the same exemplars as in the implementation of ToolkenGPT and \textsc{Tecton}. Note that Llama 3 8B Instruct achieves $79.6\%$ accuracy \cite{dubey2024llama3herdmodels} on the non-enhanced version of GSM8K when prompted in few-shot CoT fashion. Our experiments show that it performs significantly worse ($37.3\%$) on GSM8K-XL, highlighting the difficulty of the enhanced dataset.

\paragraph{\textsc{Trice}.}  \textsc{Trice} is a two-stage system where instruction tuning is followed by reinforcement learning from environmental feedback (RLEF), leveraging LoRA at both stages. We train \textsc{Trice} using the same hyperparameters as in the original implementation, with the exception of the batch size in the first phase of training, reduced to 128 with 8 gradient accumulation steps due to the memory limitations of our hardware. Since we train on a single dataset (GSM8K-XL), our implementation corresponds to the setup referred to as \textsc{Trice-split} in the original paper. Note that \citet{qiao2024trice} implement \textsc{Trice} on top of Alpaca \cite{alpacareport}, ChatGLM \cite{glm2024chatglmfamilylargelanguage} and Vicuna \cite{vicunareport}. For comparability with \textsc{Tecton} and the other baselines, here we use Llama 3 as the base model. We thus adjust \textsc{Trice}'s prompt templates and special tokens to be consistent with Llama 3's model card\footnote{{https://www.llama.com/docs/model-cards-and-prompt-formats/meta-llama-3}}.

\paragraph{ToolkenGPT.} Like \textsc{Tecton}, ToolkenGPT augments the
output matrix of an LLM with additional special tokens, each representing a tool. Only the additional tokens are tuned while the rest of the weights remain frozen.We implement ToolkenGPT with Llama 3 and train it on the same annotated datasets as we train \textsc{Tecton}, using the same sets of tools and the same hyperparameter combinations for direct comparability.

\begin{table}[t!]
\centering
\renewcommand\arraystretch{1.5}
\scalebox{0.80}{
\begin{tabular}{lcc}
\hline
\toprule
\textbf{Test set} & \textbf{Original} & \textbf{XL version} \\
\hline

\textbf{ASDiv} & {618} & {360} \\

\textbf{MAWPS} & {505} & {416} \\

\textbf{SVAMP} & {299} & {270} \\

\bottomrule
\hline
\end{tabular}
}
\caption{Test set sizes of our OOD math reasoning datasets, in their original version and in the XL version enhanced via an automated process.}
\label{tab:datasizes}
\end{table}

\section{Out-of-distribution Datasets}
\label{sec:datasets}

We enhance the test sets of ASDiv, MAWPS and SVAMP and obtain `XL' versions of each. Our goal is to replace the numbers in each test sample with larger-magnitude ones in an automated manner, yet ensuring that the new numbers in each question correctly map to a new numerical answer. To this end, we use datasets that have been annotated with the golden sequence of operations. Since the original versions of ASDiv and SVAMP do not contain such annotations, we use the versions that are part of the Lila benchmark \cite{Mishra2022Lila}. 

Firstly, we extract all the numbers from the chain of operations using regular expressions. Secondly, we search for these numbers in the question text (we use the \texttt{num2words} library\footnote{https://github.com/savoirfairelinux/num2words}  to include numbers expressed as words) and replace them with random numbers in the interval $[-10^5, 10^5]$. We replace the numbers in both the text and the annotated chain of operations. Note that we replace negative numbers with negative numbers and positive numbers with positive numbers. We also take care to substitute integers with integers and floats with floats. If we cannot match any of the numbers in the operations to the text, we discard that sample. Lastly, we evaluate the new chain of operations with the substituted numbers and store the result as the new golden answer for that data point.

This process results in datasets containing questions with the same structure and wording as the original ones, but with the numbers greatly magnified. This provides a real challenge for contemporary large language models, which can otherwise easily solve small-number arithmetic.

As our automated process discards any data samples that it is unable to process, our final test sets are typically smaller than their original counterparts. Table~\ref{tab:datasizes} reports the sizes of the original and modified test sets.

\section{Prompts}
\label{sec:prompts}
\prompt{prompt:tecton_score_choice}{\textsc{Tecton-score} meta-reasoning prompt}{In this task, you must select the correct and most plausible continuation for the answer in the text below. You should use your common sense and logical and mathematical abilities. You should directly answer by choosing the correct option. Output only the letter corresponding to the correct option.\newline\newline
[FIXED EXEMPLARS]\newline\newline
Text to continue:\newline
Question: [QUESTION]\newline
Answer: [PARTIAL ANSWER]\newline\newline
Possible answer continuations:\newline
[OPTIONS]\newline\newline
The correct continuation is: }

\prompt{prompt:tecton_generate_hints}{\textsc{Tecton-generate} meta-reasoning prompt}{Complete the answers below. In square brackets you will find some hints showing the possible math operations. You may use one of these hints if you think it is correct.\newline\newline
[DYNAMIC EXEMPLARS]\newline\newline
Question: [QUESTION]\newline
Answer: [HINTS] [PARTIAL ANSWER]}
\vspace{-8pt}
In the reasoning phase, we prompt \textsc{Tecton} to generate an answer to a math reasoning question step by step, aided by in-context exemplars. Prompts \ref{prompt:tecton_score_choice} and \ref{prompt:tecton_generate_hints} illustrate how we elicit tool choice in the meta-reasoning phase, for \textsc{Tecton-score} and \textsc{Tecton-generate} respectively. In both cases, we show the model in-context exemplars, followed by the current question and the current partial answer. The latter consists of the lines of text generated during the previous iterations of \textsc{Tecton}, if any. In Prompt~\ref{prompt:tecton_score_choice}, the placeholder \texttt{[FIXED EXEMPLARS]} is replaced with six in-context exemplars extracted from the training set. The number of options in each exemplar matches the number of available options in the current sample. In Prompt \ref{prompt:tecton_generate_hints}, we have \texttt{[DYNAMIC EXEMPLARS]} that are retrieved to demonstrate the tools currently in the candidate set. Here, the partial answer also includes the tokens generated during the current iteration, up to the sequence position where the first tool is found among the $k$ top probability tokens. The hints are prepended to the partial answer in this setting.

\section{Efficiency}

We compare the efficiency of our method with the fine-tuned baselines---\textsc{Trice} and ToolkenGPT.

\subsection{Comparison with \textsc{Trice}}
\textsc{Tecton} is more efficient than \textsc{Trice}, as it only requires tuning additional embeddings in the output layer, with minimal backpropagation. In contrast, \textsc{Trice} performs two rounds of finetuning each time updating LoRA modules throughout the model. 

\subsection{Comparison with ToolkenGPT}
We train \textsc{Tecton} in the same way as the ToolkenGPT baseline, by only updating the additional token embeddings in the output layer. At inference, \textsc{Tecton} requires generating additional text for each line in the problem. Note that for \textsc{Tecton}-\textsc{score} this overhead is negligible, as we generate only one additional token per line.

\end{document}